\documentclass[10pt,twocolumn,letterpaper]{article}

\usepackage{iccv}
\usepackage{times}
\usepackage{epsfig}
\usepackage{graphicx}
\usepackage{amsmath}
\usepackage{amssymb}

\usepackage{float}
\usepackage{subcaption}
\usepackage[linesnumbered, ruled, vlined]{algorithm2e}
\usepackage[export]{adjustbox}

\usepackage[pagebackref=true,breaklinks=true,letterpaper=true,colorlinks,bookmarks=false]{hyperref}

\iccvfinalcopy % *** Uncomment this line for the final submission

\def\iccvPaperID{2810} % *** Enter the ICCV Paper ID here

\ificcvfinal\fi
\begin{document}

%%%%%%%%% TITLE
\title{Unsupervised Domain Adaptation using Deep Networks with Cross-Grafted Stacks} %{Deep Cross-Grafted Stacks to Adapt between Domains}

%	\author{Jinyong Hou\\
%		Department of Information Science, University of Otago\\
%		%Institution1 address\\
%		{\tt\small robert.hou@postgrad.otago.ac.nz}
%		% For a paper whose authors are all at the same institution,
%		% omit the following lines up until the closing ``}''.
%		% Additional authors and addresses can be added with ``\and'',
%		% just like the second author.
%		% To save space, use either the email address or home page, not both
%		\and
%		Xuejie ding\\
%		Institute of Information Scien, Chinese Academy of Sciences\\
%		%First line of institution2 address\\
%		{\tt\small dingxuejie@iie.ac.cn}
%		\and
%		Jeremiah Deng\\
%		Department of Information Science, University of Otago\\
%		%Institution1 address\\
%		{\tt\small jeremiah.deng@otago.ac.nz}
%	}

\author{Jinyong Hou\textsuperscript{1}, Xuejie Ding\textsuperscript{2}, Jeremiah D. Deng\textsuperscript{1}, Stephen Cranefield\textsuperscript{1}\\
	%    	Department of Information Science, University of Otago\\
	%    	{\tt\small robert.hou@postgrad.otago.ac.nz}
	%    	\and
	%%    	Xuejie ding\\
	\small Department of Information Science, University of Otago\textsuperscript{1} \\
	\small Institute of Information Engineering, Chinese Academy of Sciences\textsuperscript{2} \\
	%    	%First line of institution2 address\\
	%    	{\tt\small dingxuejie@iie.ac.cn}
	%    	\and
	%    	Jeremiah Deng\\
	%    	Department of Information Science, University of Otago\\
	{\tt\small robert.hou@postgrad.otago.ac.nz,dingxuejie@iie.ac.cn,}\\
		{\tt\small \{jeremiah.deng,stephen.cranefield\}@otago.ac.nz}
}

\maketitle
%\thispagestyle{empty}

%%%%%%%%% ABSTRACT
\begin{abstract}

Current deep domain adaptation methods used in computer vision have mainly focused on learning discriminative and domain-invariant features across different domains. In this paper, we present a novel approach that bridges the domain gap by projecting the source and target domains into a common association space through an unsupervised ``cross-grafted representation stacking'' (CGRS) mechanism. Specifically, we construct variational auto-encoders (VAE) for the two domains, and form bidirectional associations by cross-grafting the VAEs' decoder stacks. Furthermore, generative adversarial networks (GAN) are employed for label alignment (LA), mapping the target domain data to the known label space of the source domain. The overall adaptation process hence consists of three phases: feature representation learning by VAEs, association generation, and association label alignment by GANs. Experimental results demonstrate that our CGRS-LA approach outperforms the state-of-the-art on a number of unsupervised domain adaptation benchmarks.
    
%	Popular deep domain adaptation methods have mainly focused on learning discriminative and domain-invariant features of different domains. In this work, we present a novel approach to bridge the domains gap by projecting the source and target into a common association space. First, representations of the source and target domains are obtained by the variational auto-encoder (VAEs) respectively. Then we construct networks with cross-grafted representation stacks (CGRS). There, it recruits the different level unsupervised representations learned by sliced receptive field, which projects the self-domain latent encodings to a new association space. Finally, we employ the generative adversarial networks (GAN) to pull the associations from the target to the source, mapped to the known label space. This adaptation process contains three phases, information encoding, association generation, and label alignment. Experimental results demonstrate the CGRS bridges the domain gap well, and the proposed model outperforms the state-of-the-art on a number of unsupervised domain adaptation scenarios.
\end{abstract}

%%%%%%%%% BODY TEXT
\section{Introduction}
In machine learning, domain adaptation aims to transfer knowledge learned previously from one or more ``source'' tasks to a new but related ``target'' domain. As a special form of transfer learning, it helps to overcome the lack of
labelled data in computer vision tasks by utilizing labelled data of 
the source domain and trying to automatically annotate unlabelled data in the target domain~\cite{Pan2010a}. 
%It helps overcome the lacking of labeled data to train a new model in computer vision applications by annotating synthetic and related data automatically. 
It may also be used to recognize unfamiliar objects in a dynamically changing environment in robotics. Therefore, in recent years domain adaptation, especially unsupervised domain adaptation, has become an appealing research topic~\cite{%Pan2010a,
Ben-David2010a,Adel2015a,Long2014a,Gopalan2011a,Yosinski2014a,Sener2016a,Herath2017a}. 

For domain adaptation to occur, it is assumed that the source and target domains are located in the same 
label space, but there is a domain bias. %For the unsupervised scenario, there are labeled data in the source domain and unlabeled data in the target domain. 
The challenge is to extract the domain-invariant representations from the data,
and find an effective mechanism to overcome the domain bias and map the unlabelled targets to the label space.
% and align their label space. 
%Then a metric learned by the source can be used to distinguish the unlabeled targets. 

To address the challenge, we propose to recruit different levels of deep unsupervised receptive fields 
from both the source and target domains and construct grafted representations for domain adaptation. Our approach 
is inspired by UNIT~\cite{Liu2017a}, but we generate the cross-domain association differently, employing
grafted deep network layers.  
%% TODO limitations of UNIT: it only does what....
%our proposed model projects the source and target into an unsupervised common association space and bridge the domain gap in an end-to-end way. %This makes the adaptation more effective and achieves a balanced performance from both direction for a scenario. 
Specifically, we construct two parallel variational auto-encoders (VAEs)~\cite{Kingma2013a} to extract the latent encodings of the source and target. Then we recruit the different parts of the decoders to construct some cross-grafted representation stacks (CGRS), which produces bi-directional cross association between 
the two domains. 
Furthermore, generative adversarial networks (GANs)~\cite{Goodfellow2014a} are employed to carry out label 
alignment (LA), so that associations between the source and target contribute to accurate classification. 

Due to these treatments our proposed CGRS-LA framework gives a promising direction for domain adaptation. Building cross associations between the domains, feature learning is hence achieved across domains, owning reduced domain dependency and increased domain-invariance, while adversarial networks further push feature representations away from the differences between domains, contributing to robust domain adaptation performance. Also, the cross-grafting process is entirely symmetric, leading to similar performance regardless of the adaptation direction, as revealed by our experiment results. Another advantage revealed by our experiments is that the CGRS is rather transferable across different tasks, which is an attractive trait for developing practical applications. %offers a number of advantages: (i) The model structure is flexible to adjust for scenarios, and decoupled from task-specific adaptation. When the associations are achieved, we can use a decoupled classifier to classify the source and target.  %In addition, it is flexible to adjust the CGRS structure according to the scenarios. 
%(ii) It has a better transferability for unseen domains. It performs well to project the objects of unseen domains to the association space for the further adaptation. (iii) It increases the robustness of adaptation and improves the performance of adaptation from both direction in scenarios. 

%In the remainder of the paper, we will briefly review some related work in Section 2. And then we outline the overall structure of our proposed model, introduce the CGRS scheme, %associations in a generative approach, 
%and present the learning metrics used by the model in Section 3. Finally, experimental results are then presented
%to demonstrate the effectiveness of our proposed model. %Empirical results demonstrate that the proposed model outperforms the state-of-the-art on various domain adaptation scenarios.

The rest of paper is organized as follows. In Section 2, we will briefly review some related work. In Section 3, we outline the overall structure of our proposed model, introduce the CGRS scheme, and present the learning metrics used by the model. Finally the experimental results are presented in Section 4. We conclude the paper in Section 5, indicating our plan of future work. %Empirical results demonstrate that the proposed model outperforms the state-of-the-art on various domain adaptation scenarios.     

\section{Related Work}
%% A BIT OF OUT OF PLACE - moved after 'intermediate representations'

%In our proposed model, the CGRS projects the self-domain latent encodings to a same intermediate association space firstly. 
There are existing works that utilize intermediate feature representations to transfer previously learned knowledge to the target tasks. Self-taught learning~\cite{Raina2007a} uses unsupervised learning trained on natural images to construct a sparse coding space, to which targets are projected to complete the recognition. In geodesic flow kernel~\cite{Gong2017,Gopalan2014}, the source and target datasets are embedded in a Grassman manifold, and a geodesic flow is constructed between the domains. A number of feature subspaces are sampled along the geodesic flow, and a kernel can be defined on the incremental feature vector, allowing a classifier to be built for the target dataset. DLID~\cite{Chopra2013a} uses deep sparse learning to extract the interpolated representation from a set of intermediate datasets constructed by combining the source and target datasets using progressively varying proportions, and the features from these intermediate datasets are concatenated to train a classifier. %Contrast with exiting works, our CGRS is learned from source and target to build the connection between them. And it is decoupled and flexible according to the different ratios of the recruit. On the other hand, the CGRS is generative, the visible intermediate associations bring a better understanding of the adaptation. 

%Then we adopt the adversarial strategy to confuse the domains.
Recent works have shown that deep networks involved in domain adaptation have achieved impressive performance due to their strong feature learning capacity. This provides a considerable improvement for some cross-domain recognition tasks~\cite{Venkateswara2017a,Long2015a,Tzeng2015a,Long2016a,Rozantsev2018a, Liu2017a,Bousmalis2016a,Ghifary2016a}.
Specifically, a number of deep domain adaptation models have applied the adversarial training strategy~\cite{Tzeng2017a,Tzeng2014a,Ganin2016a,Liu2017a,Bousmalis2017a,Liu2016a,Liu2017b}. DANN~\cite{Ganin2016a} employs a gradient reversal layer between the feature layer and the domain discriminator,
causing feature representation to anti-learn the domain difference and hence adapt well to the target domain. %During the training, this confused the domain discriminator, and adapted the features of target to the classifier trained by source. 
ADDA~\cite{Tzeng2017a} firstly trains a convolutional neural network (CNN) using the source dataset. An adversarial phase then follows, with the CNN assigned to the target for domain discriminator training, and the new target encoder CNN is finally combined with source classifier to achieve the adaptation. %In addition, the authors presented a general framework for the adversarial domain adaptation.  

Using generative adversarial networks (GAN), the PixelDA framework~\cite{Bousmalis2017a} generates synthetic images from source-domain images that are mapped to the target domain. A task classifier then is trained by the source and synthetic images using the source labels. UNIT~\cite{Liu2017a} introduces an unsupervised image-to-image translation framework based on couple of variational auto-encoders (VAEs) and GANs. %UNIT aims at learning a joint distribution of images in different domains by using images from the marginal distributions in individual domains. 
To achieve this, a pair of corresponding images in different domains are mapped to a shared latent representation space. 

Inspired by these previous works, our proposed CGRS-LA framework combines two ideas: constructing cross-domain feature representations, and employing adversarial networks for label alignment. Specifically, it incorporates VAEs to learn feature representations, a cross-grafting step to generate bidirectional cross-domain associations, and a generative adversarial approach that carries out classification on source-target associations. A detailed descriptions of our framework are given next. %, which is between the associations rather than from the source to the target directly. This makes the adversarial process soft and effective. 

%%figure for the model description
\begin{figure*}
	\centering
	\includegraphics[width=0.6\textwidth]{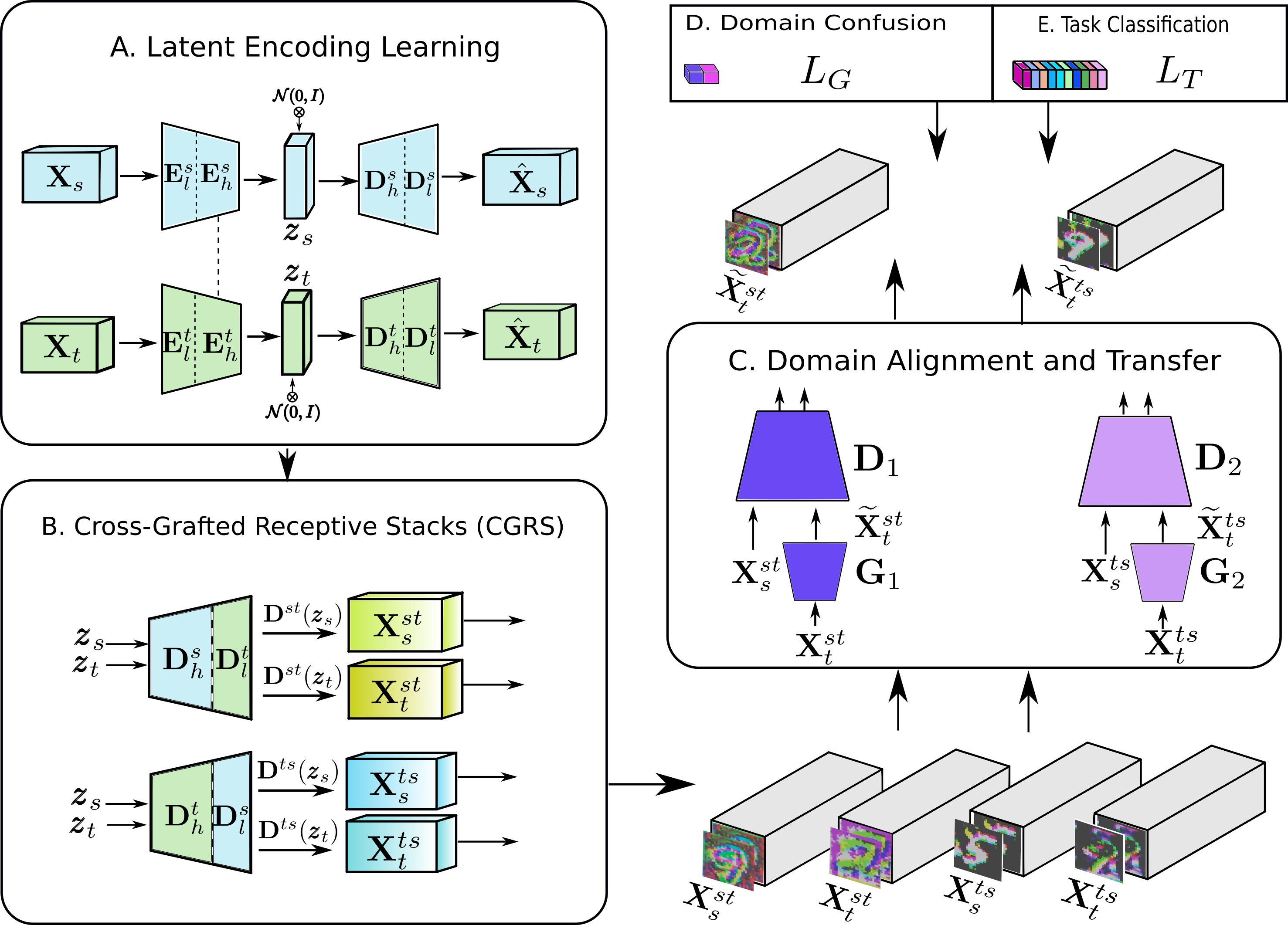}
	\caption{\small Overview of the the proposed model. There are 5 modules in it. In module \textit{A}, the high-level layers of encoders $E_{h}^{s}$, $E_{h}^{t}$ are shared (demonstrated by the dashed line).  The outputs of $D_{h}^{s}$ and $D_{h}^{t}$ are the high-level representation of the source and target, whereas $D_{l}^{s}$, $D_{l}^{t}$ are the low-level ones. The  $\mathbf{X}_s^{st}$, $\mathbf{X}_t^{ts}$, $\mathbf{X}_s^{ts}$, $\mathbf{X}_t^{ts}$ in module \textit{B} are the association images reproduced by CGRS ($D^{st}\equiv \lbrack D_{h}^{s}\circ D_{l}^{t}\rbrack$ and $D^{ts}\equiv \lbrack D_{h}^{t}\circ D_{l}^{s}\rbrack$) from latent encodings. In module \textit{C}, $G_1$ and  $G_2$ are adversarial generators, $D_1$, $D_2$ are discriminators. $L_{G}$ and $L_{T}$ are learning metric for the domain and task respectively. Best viewed in color.}
	\label{fig:model_schem}
\end{figure*} 

%--------------------------------------------------------------------------------------------------------------------------------------

\section{The CGRS-LA Framework}
\subsection{Model Description} 

%For the domain adaptation in computer vision applications, 
We consider two domains: one is a source domain $\mathcal{D}_s$, which is constructed by $n_s$ images $\mathbf{X}_s=\{\pmb{x}_{i}^{s}\}_{i=1}^{n_s}$ and their correspond labels $\pmb{y}_s=\{y_i^s\}_{i=1}^{n_s}$; the other is a target domain $\mathcal{D}_t=\{\mathbf{X}_t, \pmb{y}_t\}$, where $\mathbf{X}_t=\{\pmb{x}_{i}^{t}\}_{i=1}^{n_t}$ and their labels $\pmb{y}_t=\{y_i^t\}_{i=1}^{n_t}$ are not available during adaptation. The source and target domain are drawn from joint distributions $\mathcal{P}(\mathbf{X}_s, \pmb{y}_s)$ and $\mathcal{Q}(\mathbf{X}_t, \pmb{y}_t)$, with a domain bias making $\mathcal{P}$ and $\mathcal{Q}$ different. Our goal is to learn some representations bearing similarity to both domains, i.e.~some joint distribution between $\mathcal{P}$ and $\mathcal{Q}$ as a bridge for knowledge transfer, based on which the target images can be successfully classified.

Our framework is shown in Figure~\ref{fig:model_schem}, split into five modular sub-tasks based on the ideas outlined as above. Firstly, in module \textit{A}, the VAEs couple are implemented by CNNs. Both the encoders and decoders are divided into high and low level stacks. The high-level layers of the encoders are shared between domains. The source and target data are encoded to latent representation $\pmb{z}_{s}$ and $\pmb{z}_{t}$, and then decoded to the reconstruction images $\widehat{\pmb{x}}_{s}$ and $\widehat{\pmb{x}}_{t}$ respectively. We assume that they have the same latent space, and the prior distribution is a normal one, $\mathcal{N}(0, I)$. 

Secondly, the latent encodings pass through the cross-grafted stacks, forming cross-domain associations that are aligned to the label space. 
%, which includes CGRS and domain alignment module. 
In module \textit{B}, we construct two parallel CGRS by grafting the decoder stacks of the source and the target. Therefore, the cross-domain association images ($ {\mathbf{X}_s^{st}},{\mathbf{X}_t^{st}},{\mathbf{X}_s^{ts}},{\mathbf{X}_t^{ts}})$ are generated when the latent encodings from different domains (indicated by subscripts) pass through the CGRS (order indicated by superscripts). The detailed generation of associations is described in the next section. In the domain alignment module \textit{C}, $G_1$ and $G_2$ are two adversarial generators for associations. They are used to generate the target association adversarial to the source's association, and vice versa. The situation when the source associations works as the ``real player'' for the adversarial generation is shown in Figure~\ref{fig:model_schem}\footnote{The arrangement can be flexible, i.e. it also works if the target association is used as the real player.}. Here the adversarials of the corresponding target associations are $\widetilde{\mathbf{X}}_t^{st}$, and $\widetilde{\mathbf{X}}_t^{ts}$. The discriminators $D_1$, $D_2$ are used to distinguish associations of  ${\mathbf{X}_s^{st}}$ from $\widetilde{\mathbf{X}}_t^{st}$, and ${\mathbf{X}_s^{ts}}$ from $\widetilde{\mathbf{X}}_t^{ts}$ respectively.  %, $\lbrack D_{sh}\circ D_{tl}\rbrack$ and $\lbrack D_{th}\circ D_{sl}\rbrack$

Finally, $L_{G}$ and $L_{T}$ in module \textit{D} and \textit{E} are the learning metrics for domain confusion and task classification. Module \textit{C} combines the learning metric modules to align the label space of the source and target images, and complete the adaptation. The training process adopts standard back-propagation. In contrast to the conventional domain adaptation framework in which the classifier input is $\{ \mathbf{X}_s, \pmb{y}_s\}$ and output is $\{ \mathbf{X}_t,\widehat{\pmb{y}_t}\}$, our model's classifier is trained by $\{\mathbf{X}_s^{st},\pmb{y}_s\}$, $\{\mathbf{X}_s^{ts},\pmb{y}_s\}$ and tested by $\{ \widetilde{\mathbf{X}}_t^{st},\pmb{y}_t\}$, $\{ \widetilde{\mathbf{X}}_t^{ts},\pmb{y}_t\}$. In short, the associations of the source data are used for training, and the adversarial generation of the target data are used in  testing. 

%---------------------------------------------------------------------------
\subsection{Generation of Cross-Grafted Associations}
%% figure for the section 3.2 description for association
\begin{figure*}
	\centering
	\includegraphics[width=0.7\textwidth]{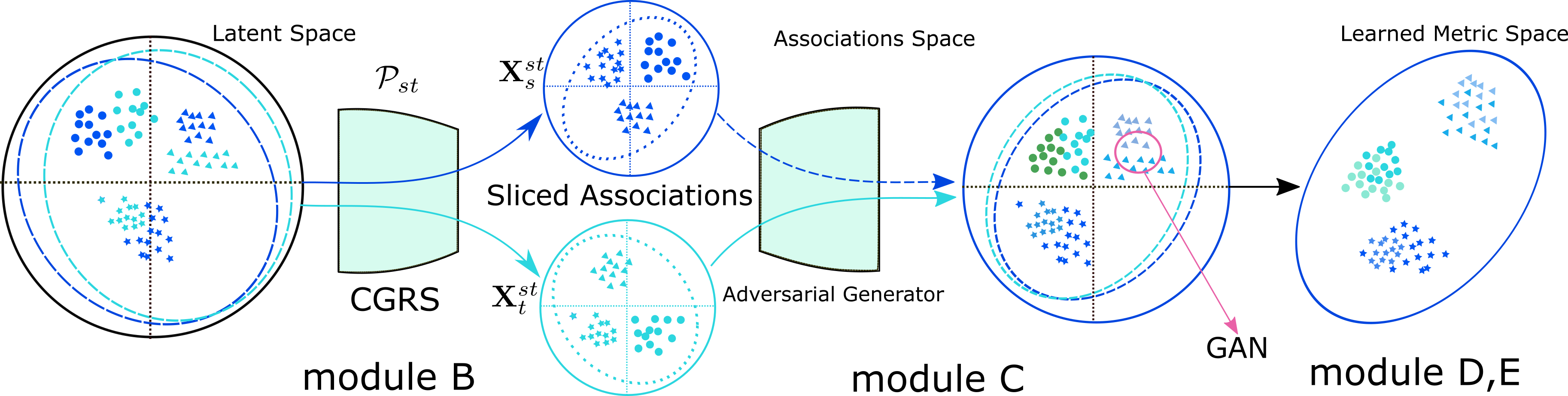}
	\caption{\small Generation of associations for channel $\mathbf{X}^{st}$. The encodings of source $s$ and target $t$ are formed in the latent space first. Then, they are projected to association spaces by CGRS. Finally, the latent and association spaces are aligned by the adversarial training combined with learning metrics. The adversarial process is flexible, and can be from the target to the source, and vice versa. The former scenario is shown here. The dashed line means the source associations are the real player in adversarial generation.}
	\label{fig:associations}
\end{figure*}

In module \textit{A}, we obtain the latent encodings of source and target domains using VAEs~\cite{Kingma2013a}, assuming they have a normal prior distribution. They encode a data sample $\pmb{x}$ to a latent space $\pmb{z}$ and decode the latent representation back to data space image $\widehat{\pmb{x}}$. We get all the latent encodings $\pmb{z}_s$ and $\pmb{z}_t$, which are conceptually sampled from conditional probability densities $q(\pmb{z}_s|\mathbf{X}_s)$ and $q(\pmb{z}_t|\mathbf{X}_t)$ respectively. %Figure~\ref{fig:model_schem} have shown that there are two modules to generate the association space. 
In module B, the cross-grafted receptive stacks are constructed to map the encodings to the cross-domain association spaces, which are later aligned to the source domain's label space in module C. %And in module \textit{C}, we try to align the domains according to the associations. 

Here CGRS recruits the high level ($i$) and low level ($j$) of the decoders of source ($s$) and target ($t$). It maps the latent space $\pmb{z}_{k}$ to the common association distributions $\mathcal{P}_{ij}$, which the associations $\mathbf{X}_{k}^{ij}$ are sampled from: 
\begin{equation}\label{eq:x_to_rho}
\textstyle D_{CGRS}(\pmb{z}_{k})\mapsto \mathbf{X}_{k}^{ij} \in \mathcal{P}_{ij},
\end{equation}
where $i,j,k \in \{s,t\}, i \neq j $. In detail, when the latent encoding $\pmb{z}_{k}$ passes through CGRS, the generation of associations can be expressed in a generative approach~\cite{Bengio2013a} as follows: 
\begin{equation}\label{eq:high_level_associations}
\textstyle \mathcal{P}_i=\textstyle (\prod_{l^{'}=1}^{N}p_{i}(\pmb {m}^{l^{'}+1}| \pmb{m}^{l^{'}},\theta_{D_{ih}}^{l^{'}}))~p_{i}(\pmb {m}^1|\pmb{z}_{k},\theta_{D_{ih}}^{1}),
\end{equation}
where $N$ is the number of high-level decoder layers, $\theta_{D_{ih}}^{l^{'}}$ is the map parameters of $i$ in $l^{'}$ layer, $\pmb{m}^{l^{'}}$ is the output space of high-level decoder of layer $l^{'}$. Then $\pmb{m}^N$ is transferred to final association space:% further as follows:   % In our paper, the transpose convolution is used to achieve the output space. with the tied structure of encoders 
\begin{equation}\label{eq:low_level_associations}
\textstyle \mathcal{P}_{ij}=\textstyle (\prod_{l^{''}=1}^{M}p_{j}(\pmb {n}^{l^{''}+1}| \pmb{m}^{l^{''}},\theta_{D_{jl}}^{l^{''}}))~p_{j}(\pmb {n}^1|\pmb{m}^{N},\theta_{D_{jl}}^{1}),
\end{equation}
where $M$ is the number of low-level decoder layers, $\pmb{n}^{l^{''}}$ is the output space of low-level decoder of layer $l^{''}$, and $\theta_{D_{jl}}^{l^{''}}$ is the map parameters of $j$ in $l^{''}$ layer. We assume the grafted parts $p_{j}$ can be regarded as the corresponding reconstruction decoder injected with noise $\epsilon_{jl}^{l^{''}}$, $\epsilon_{jl}^{l^{''}}\in \mathcal{P}(\theta_{D_{jl}}^{l^{''}}|\mathbf{X}_{j})$ in a normal distribution (more details in the supplementary). This bridges the gap between the source and target domains, and also enhances the generalization of the model. Figure~\ref{fig:associations} gives the schematic illustration for the generation of associations $\mathbf{X}_{s}^{st}$ and $\mathbf{X}_{t}^{st}$, when $i=s, j=t$. Another case is $i=t, j=s$, which corresponds to the associations $\mathbf{X}_{s}^{ts}$ and $\mathbf{X}_{t}^{ts}$. 

The associations are constructed, but they are yet to be aligned to the same label space of the source domain. To get the label distributions aligned, we use discriminator \textit{D} to confuse the associations generated by different encodings. The discriminators make the distributions of associations more similar by minimizing the Jensen-Shannon divergence~\cite{Goodfellow2014a} ($JSD$):% and can be expressed as follows,
\begin{equation}\label{eq:jsd1}
\begin{gathered}
%\textstyle p(\mathbf{X}_{s}^{st}|\pmb {z}_{s}, \theta_E, \theta_D) \Leftarrow p(\mathbf{X}_{t}^{st}|\pmb {z}_{t}, \theta_E, \theta_D, \theta_G) \\
\textstyle \widetilde{\mathbf{X}}_t^{ij} \in p(\widetilde{\mathbf{X}}_t^{ij}|\mathbf{X}_t^{ij},\theta_D,\theta_{G_{k}}) \\
\textstyle w.r.t \  \min JSD (p(\mathbf{X}_{s}^{ij})\|p(\widetilde{\mathbf{X}}_{t}^{ij})),
\end{gathered}
\end{equation}
%\begin{equation}\label{eq:jsd2}
%\begin{gathered}
%\textstyle p(\mathbf{X}_{s}^{ts}|\pmb {z}_{s}, \theta_E, \theta_D) \Leftarrow p(\mathbf{X}_{t}^{ts}|\pmb {z}_{t}, \theta_E, \theta_D, \theta_G)  \\
%\textstyle w.r.t \  \min JSD (p(\mathbf{X}_{s}^{ts})\|p(\mathbf{X}_{t}^{ts})).
%\end{gathered}
%\end{equation}
In the model, $\theta_{G_{k}}$ ($G_{1}$ when $i=s, j=t$ and $G_{2}$ for $i=t, j=s$) are used as generators for $\widetilde{\mathbf{X}}_{t}^{st}$ and $\widetilde{\mathbf{X}}_{t}^{ts}$ during the alignment, as shown in Figure~\ref{fig:model_schem}. The encoders in module \textit{A} and adversarial generators of module \textit{C} are updated during training to minimum the Jensen-Shannon divergence of associations.

%% algorithm of the model

%---------------------------------------------------------------------------
\subsection{Learning}

To train our model, we jointly solve the learning problems of the subnetworks. There are four loss functions, for the within-domain VAEs~\cite{Kingma2013a}, cross-domain adversarial networks, content constancy and classifier training loss respectively.

First, we need to learn the representations of the source and target domains from encoders and decoders. Here, we minimize the within-domain VAEs loss functions. The loss function of our VAEs consists of both reconstruction error and prior regularization: % two parts. The loss function is:
\begin{equation} \label{eq:overall_vae}
L_{VAEs}=L_{like}^{pixel}+L_{prior}.
\end{equation}
The $L_{like}^{pixel}$ and $L_{prior}$ are given by
\begin{equation} \label{eq:vae_rec}
\begin{split}
\textstyle L_{like}^{pixel}  &= \textstyle -\lambda_1\{ \mathbb{E}_{q_{s}(\pmb{z}_s|\mathbf{X}_s)}[\log p_{s}(\mathbf{X}_s|\pmb{z}_s)] \\
\textstyle &+ \mathbb{E}_{q_{t}(\pmb{z}_t|\mathbf{X}_t)}[\log p_{t}(\mathbf{X}_t|\pmb{z}_t)]\} ,\\
\end{split}
\end{equation}
\begin{equation} \label{eq:vae_latent}
\begin{split}
\textstyle L_{prior} &= \textstyle \lambda_2\{ D_{KL}(q_{s}(\pmb{z}_s|x_{s})||p(z)) \\
\textstyle &+ D_{KL}(q_{t}(\pmb{z}_t|x_{t})||p(\pmb{z}))\},\\
\end{split}
\end{equation}
where $D_{KL}$ is the Kullback-Leibler divergence. $\lambda_1$ and $\lambda_2$ are the trade-off hyper-parameters to control the priority of variational encoding and reconstruction. 

To align the source and target domains, we use the adversarial training for the two association spaces $\mathcal{P}_{st}$ and $\mathcal{P}_{ts}$. %During the experiments, 
Their adversarial objectives $L_G^{st}$ and $L_G^{ts}$ are: 
\begin{equation} \label{eq:gan_st}
\begin{split}
\textstyle L_G^{st}(E_s,D^{st},D_1) &= \textstyle \lambda_0\{ \mathbb{E}_{x_{s}}[\log D_{1}(D^{st}(\pmb{z}_s))] \\
\textstyle &+ \mathbb{E}_{x_{s},z_{s}}[\log (1- D_{1}(G_1(D^{st}(\pmb{z}_t))))]\},
\end{split}
\end{equation}
\begin{equation} \label{eq:gan_ts}
\begin{split}
\textstyle L_G^{ts}(E_t,D^{ts},D_2) &=\textstyle  \lambda_0\{ \mathbb{E}_{x_{t}}[\log D_{2}(D^{ts}(\pmb{z}_s))] \\
\textstyle &+ \mathbb{E}_{x_{t},z_{t}}[\log (1- D_{2}(G_2(D^{ts}(\pmb{z}_t))))]\},
\end{split}
\end{equation}
where $D^{st}\equiv D_{h}^{s}\circ D_{l}^{t}$ and $D^{ts}\equiv D_{h}^{t}\circ D_{l}^{s}$. $D(x)$ is the probability function assigned by the discriminator network, which tries to distinguish the generated source-based associations from the target-based ones. At last, the overall adversarial generative cost function is:
\begin{equation} \label{eq:overall_gan}
\textstyle L_{G}=\textstyle L_G^{st}+L_G^{ts}.
\end{equation}
%During the experiments, 
For the training stability, we introduce a content constancy loss function for the associations. Both the $L1$ and $\mathnormal{L}2$ penalty can be used to regularize the associations. Here we render a masked pairwise mean squared error~\cite{Bousmalis2017a}. Formally, when a binary mask $\mathbf{m}$ is given ($\mathbf{m}\in\mathcal{R}^k$), the masked PMSE loss for associations $\mathbf{X}^{st}$ and $\mathbf{X}^{ts}$ is given as follows:
\begin{equation} \label{eq:pmse_st}
\begin{split}
\textstyle L_s^{st} &= \textstyle \mathbb{E}_{\mathbf{X}_s^{st}, z}(\frac{1}{k}||D^{st}(\pmb{z}_s) - G_1(D^{st}(\pmb{z}_t)) \circ \mathbf{m}||_2^2 \\
\textstyle &- \frac{1}{k^2}((D^{st}(\pmb{z}_s) - G_1(D^{st}(\pmb{z}_t)))^T\mathbf{m})^2),
\end{split} 
\end{equation}
and
\begin{equation} \label{eq:pmse_ts}
\begin{split}
\textstyle L_s^{ts} &=\textstyle \mathbb{E}_{\mathbf{X}_s^{ts}, z}(\frac{1}{k}||D^{ts}(\pmb{z}_s) - G_2(D^{ts}(\pmb{z}_t)) \circ \mathbf{m}||_2^2 \\
\textstyle &- \frac{1}{k^2}((D^{ts}(\pmb{z}_s) - G_2(D^{ts}(\pmb{z}_t)))^T\mathbf{m})^2).
\end{split} 
\end{equation}
So the overall content objective for associations is:
\begin{equation} \label{eq:overall_pmse}
\textstyle L_s = \lambda_3(L_s^{st} + L_s^{ts}).
\end{equation}
At last, for classification we use a typical soft-max cross-entropy loss:
\begin{equation} \label{eq:classifier}
\textstyle L_{T} = \mathbb{E}[-y_{s}^{T}\log T(\mathbf{X}_s^{st}) - y_{s}^{T}\log T(\mathbf{X}_s^{ts})],
\end{equation}
where $y_{s}$ is the class label for source $\mathbf{X}_{s}$, and $T$ is the task classifier. Finally, the overall loss function of our model is:
\begin{equation} \label{eq:overall_loss_function}
L^*%_{overall}
=\mathop{\min}\limits_{E, D, G}\mathop{\max}\limits_{D_1, D_2}(L_{VAEs}+ L_{G}+L_{s}+L_{T}).
\end{equation}
We solve this minimax problem of the loss function optimization by three alternating steps. First, the latent encodings are learned by the self-mapped process, which updates $(E_s, E_t, D_s, D_t) $, but keeps CGRS $(D^{st},D^{ts})$, ($D_1, D_2$) and ($G_1, G_2$) fixed. Then, we apply a gradient ascent step to update two discriminators $D_1$, $D_2$ and the classifier $T$, while keeping two VAEs channels $(E_s,E_t,D_s,D_t) $ and CGRS $(D^{st}, D^{ts})$, ($G_1, G_2$) fixed. Finally, a gradient descent step is applied to update $(E_1, E_2, G_1, G_2)$,  while $(D^{st},D^{ts})$, $D_1$, $D_2$ and $T$ are fixed.
%%figure of datasets
\begin{figure}
	\centering
	\includegraphics[width=0.42\textwidth]{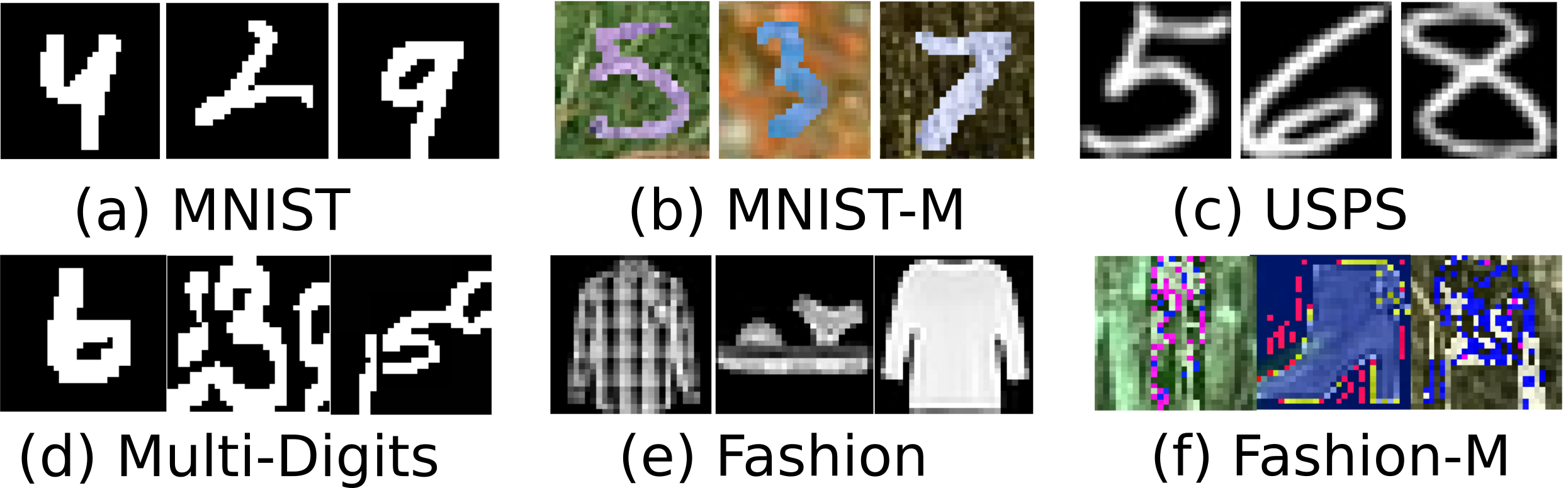}
	\caption{\small Examples of the Datasets used for Experiments.}
	\label{fig:datasets}
\end{figure}  

%--------------------------------------------------------------------------------------------------------------------------------------
\section{Experiments and Results}

We have evaluated our model on some benchmark datasets used commonly in the domain adaptation literature, including MNIST~\cite{LeCun1998a}, MNIST-M~\cite{Ganin2016a}, and USPS~\cite{LeCun1989a}. Also included is a multi-digit dataset ``M-Digits'', which we developed based on MNIST. The Fashion dataset~\cite{Xiao2017} and its polluted version ``Fashion-M'' are also used in the experiments. Example images of these datasets are shown in Figure~\ref{fig:datasets}. 

%	We consider the following unsupervised domain adaptation scenarios: (a) MNIST and MNIST-M; (b) MNIST and USPS; (c) Fashion and Fashion-M; (d) MNIST and M-Digits.

We compare our CGRS-LA method with the state-of-the-art domain adaptation methods: Pixel-level domain adaptation (PixelDA)~\cite{Bousmalis2017a}, Domain Adversarial Neural Network (DANN)~\cite{Ganin2016a}, Unsupervised Image-to-Image translation (UNIT)~\cite{Liu2017a}, Cycle-Consistent Adversarial Domain Adaption (CyCADA)~\cite{Hoffman2018a}, Generate to Adapt (GtA)~\cite{Sankaranarayanan2018a} and Conditional Domain Adversarial Network (CDAN)~\cite{Long2018a}. 
In addition, we also use the source-only and target-only training as the lower and upper bound respectively, following the practice in~\cite{Bousmalis2017a, Ganin2016a}. %For the source-only training, the model is trained on the source dataset only, and then is tested on the target dataset. When the target dataset is used to train and test, this is target-only scenario.

%---------------------------------------------------------------------------
\subsection{Datasets and Adaptation Scenarios}
We use six popular datasets to construct four domain adaptation scenarios:
%% table for acc of adaptation 
\begin{table*} 
	\centering
	\caption{\small Mean classification accuracy comparison. The "source only" row is the accuracy for target without domain adaptation training only on the source. And the "target only" is the accuracy of the full adaptation training on the target. For each source-target task the best performance is in bold. }
	\begin{tabular}{|l|c|c|c|c|c|c|c|c|}
		\hline
		Source & MNIST & MNIST-M & MNIST & USPS & MNIST  & M-Digits & Fashion & Fashion-M \\
		Target & MNIST-M & MNIST & USPS  & MNIST & M-Digits & MNIST & Fashion-M & Fashion \\
		\hline\hline
		Source Only & 0.561  &  0.633   & 0.634 & 0.625  & 0.603 & 0.651 & 0.527 & 0.612   \\
		\hline\hline
		CORAL~\cite{Sun2016a} & 0.817 & -  & 0.577 &  -   &  - &  - &  - &  -   \\
		MMD~\cite{Long2015a} & 0.811 & -   & 0.769 &  -  &  - & -  & -  &  -   \\
		CyCADA~\cite{Hoffman2018a} & - & - & 0.956  & 0.965  & - & - & - & - \\
		GtA~\cite{Sankaranarayanan2018a} & - & - & 0.953  & 0.908  & - & - & - & - \\
		CDAV~\cite{Long2018a} & - & - & 0.956 & \textbf{0.980} & - & - & - & - \\
		DANN~\cite{Ganin2016a} & 0.766 & 0.851  & 0.774 & 0.833   & 0.864  & 0.920  & 0.604  & 0.822    \\
		PixelDA~\cite{Bousmalis2017a} & 0.982 & 0.922   & 0.959 & 0.942   & 0.734 & 0.913  & 0.805   &  0.762   \\
		UNIT~\cite{Liu2017a} & 0.920 &  0.932 &  0.960 & 0.951   & 0.903  & 0.910  & 0.796   &  0.805     \\
		\hline\hline
		CGRS-LA ($\mathbf{X}^{st}$) &  0.821 & 0.935  & 0.946  & 0.938  & 0.895 & 0.902  & 0.735   & 0.805     \\
		CGRS-LA ($\mathbf{X}^{ts}$) & 0.923  &  0.840   & 0.902  &  0.930  & 0.853  & 0.851  & 0.792   & 0.760     \\
		CGRS-LA-C ($\mathbf{X}^{st}$) &  0.890 & \textbf{0.983}  & \textbf{0.961}  & 0.956  & \textbf{0.916} & \textbf{0.923}  & 0.766   & \textbf{0.825}     \\
		CGRS-LA-C ($\mathbf{X}^{ts}$) & \textbf{0.983}  &  0.871   & 0.943  &  0.953  & 0.883  & 0.892  & \textbf{0.813}   & 0.811     \\
		\hline \hline
		Target Only & 0.983  & 0.985  &0.980 &  0.985   & 0.982  & 0.985  & 0.920  &  0.942   \\
		\hline  	
	\end{tabular}
	\label{table:accuracy}
\end{table*}

\noindent
\textbf{MNIST $\rightleftarrows$ MNIST-M:} This is a scenario when the image content is the same, but the target data are polluted by noise. MNIST handwritten dataset~\cite{LeCun1998a} is a very popular machine learning dataset. It has a training set of 60,000 binary images, and a test set of 10,000. There are 10 classes in the dataset. MNIST-M~\cite{Ganin2016a} is a modified version for the MNIST, with random RGB background cropped from the Berkeley Segmentation Dataset\footnote{URL https://www2.eecs.berkeley.edu/Research/Projects/CS/vision/bsds/}. In our experiments, we use the standard split of the dataset.

\noindent
\textbf{MNIST $\rightleftarrows$ USPS:} For this scenario, source and target domains have different contents but the same background. USPS is a handwritten zip digits datasets~\cite{LeCun1989a}. It is collected by the U.S Postal Service from envelopes processed at the Buffalo, N.Y Post Office. It contains 9298 binary images ($16\times16$), 7291 of which are used as the training set, while the remaining 2007 are used as the test set. The USPS samples are resized to $28\times28$, the same as MNIST.   

\noindent
\textbf{Fashion $\rightleftarrows$ Fashion-M:} Fashion-MNIST~\cite{Xiao2017} contains 60,000 images for training, and 10,000 for testing. All the images are grayscale, $28\times28$ in size space.%, the same as MNIST. 
The samples are collected from 10 fashion categories: T-shirt/Top, Trouser, Pullover, Dress, Coat, Sandal, Shirt, Sneaker, Bag and Ankle Boot. There are some complex textures in the images. In addition, following the protocol in~\cite{Ganin2016a}, we add random noise to the Fashion images to generate the Fashion-M dataset.
%%WHY?? 
%During the experiments, we convert the gray images to the RGB.

\noindent
\textbf{MNIST $\rightleftarrows$ M-Digits} In this scenario, we design a multi-digits dataset to evaluate the proposed model, noted as M-Digits. The MNIST digits are cropped first, and then are randomly selected, combined and randomly aligned in a new image, limited to 3 digits in maximum. The label for the new image is decided by the central digit. Finally, the new dataset is resized to $28\times28$. 
%% IRRELEVANT??
%It is similar to SVHN, but there is a big gap between the digits size for the different combination compared with SVHN.      

%---------------------------------------------------------------------------
\subsection{Implementation Details}

All the models are implemented using the TensorFlow\footnote{Our code will be available on github after the double blind review.}~\cite{Abadi2016} and are trained with Mini-Batch Gradient Descent using the Adam optimizer~\cite{Kingma2014c}. The initial learning rate is 0.0002. Then it adopts an annealing method, with a decay of 0.95 after every 20,000 mini-batch steps. The mini-batch size for both the source and target domains are 64 samples, and the input images are rescaled to [-1, 1]. The hyper-parameters are $\lambda_0 = 1$, $\lambda_1=10$, $\lambda_2=0.01$, $\lambda_3=1$.   

%	\footnote{Our code is available here: \href{https://github.com/robertholding/UDAR}{github.com/robertholding/UDAR}}

In our implementation, the latent space is sampled from a normal distribution $\mathcal{N}(0, I)$, and is achieved by the convolution encoders. The transpose convolution~\cite{Zeiler2010} is used in the decoder to build the reconstruction image space. This follows a similar structure protocol of~\cite{Liu2017a}, but we modify the padding strategy to `same' for convolution layers. For sake of convenience in experiments, we add another 32-kernel layer before the last layer in the decoders. The stride is 2 for down-sampling in the encoders, and their counterpart in decoders is also 2 so as to get the same dimensionality of the original image. The encoders for source and target domains share their high-level layers. We add the batch normalization between each layer in the encoders and the decoders. The CGRS of associations is the composition of different levels of the source and target's representation. The stride step keeps 1 for all the dimensions in the adversarial generator, and the kernel is $3\times3$. This adopts the structure of PixelDA~\cite{Bousmalis2017a}, which uses a ResNet architecture. The discriminator fuses the domains, and also plays as a task classifier for the label space learning. It follows the design as in~\cite{Liu2017a}. However, we do not share the layers of discriminators of $\mathbf{X}^{st}$ and $\mathbf{X}^{ts}$ channels. Also, we replace the max-pooling with a stride of $2\times2$ steps. 

%---------------------------------------------------------------------------
\subsection{Results}
\subsubsection{Quantitative Results}

%In the practical domain adaptation scenarios, the model is used to infer the label of unlearned objects. 
Now we report the classification performance of our proposed model. 
During the experiments, associations $\mathbf{X}_s^{st}$ and $\mathbf{X}_t^{ts}$ are used to train the classifier,  and the adversarial generation of $\mathbf{X}_t^{st}$ and $\mathbf{X}_t^{ts}$ are used for testing. 
The accuracy of the target domain classification after domain adaptation is listed in Table~\ref{table:accuracy}, presenting the result of 12 methods (4 versions of our model CGRS-LA, and 8 state-of-the-art methods) across 4 tasks (each in two directions). Our proposed model outperforms the state-of-the-art in most of the scenarios, especially when content constancy is considered. Also, it can be seen that the adaptation performance is usually asymmetric for the methods in comparison, e.g.~the accuracies for MNIST$\rightarrow$M-Digits and M-Digits$\rightarrow$MNIST are quite different for DANN and PixelDA. The CGRS-LA models, however, perform almost equally well on both directions for these adaptation tasks. 
%Usually, it is not an equal task for two datasets adapted from two directions in one scenario. Our proposed model performs well on the bidirectional task of the scenarios. 

For MNIST$\rightleftarrows$MNIST-M and MNIST$\rightleftarrows$USPS, the mean classification accuracy nearly reaches the upper bound, suggesting these are easier tasks. On the other hand, we can see the adaptation task between Fashion and Fashion-M is more difficult than others. For this task, our method again not only achieves the best performance but also demonstrates balanced performance in two directions. 
%In addition, the two channels  show some differences in classification accuracy. For the transfer task between MNIST and MNIST-M, the difference of two channels is a litter more obvious about 0.1. This is due to the unsymmetrical representation learned from the source and target. 

%%  figure for visualization
\begin{figure*}
	\centering
	\begin{subfigure}[t]{.45\textwidth}
		\centering
		\includegraphics[width=0.95\textwidth]{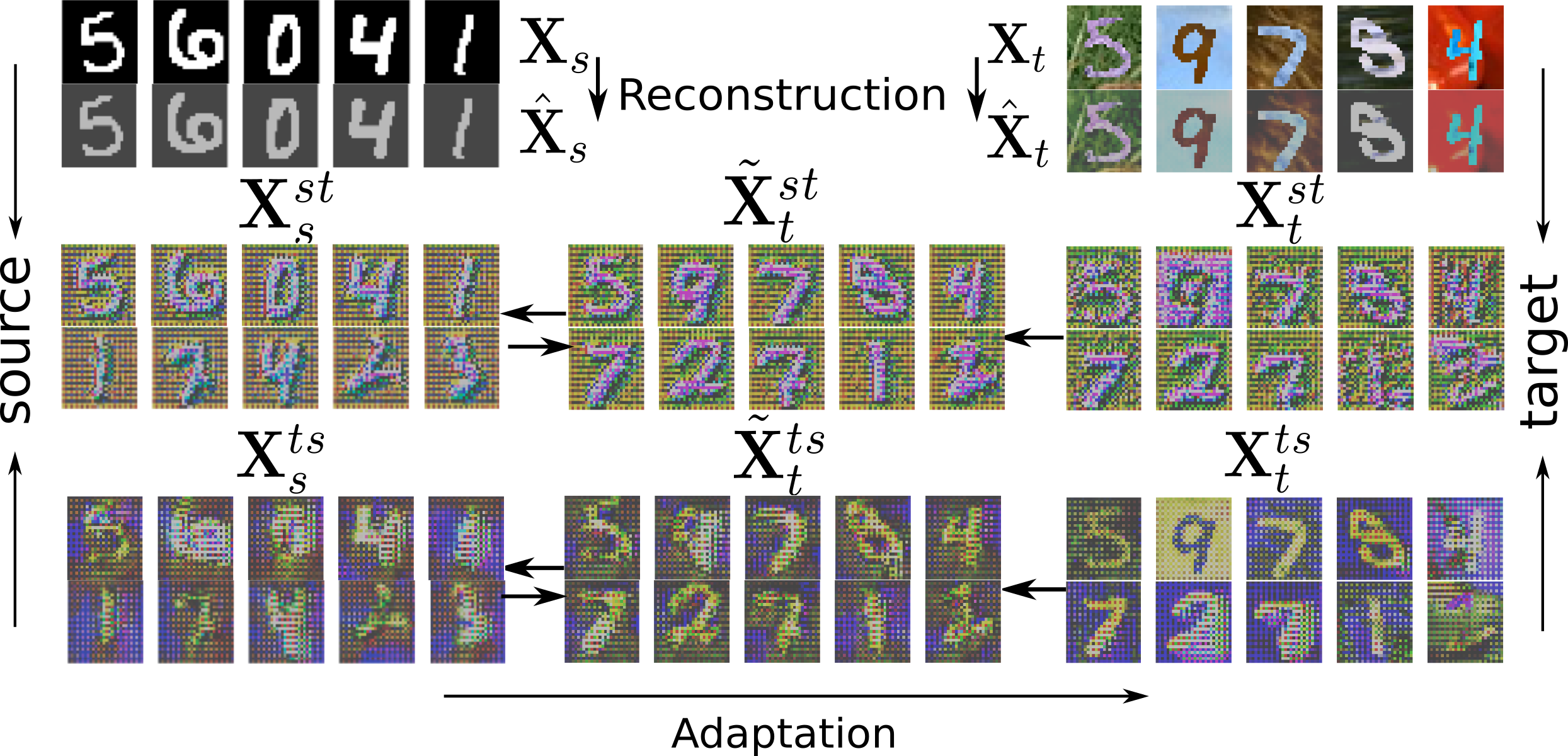}
		\caption{\small MNIST $\rightarrow$ MNIST-M}
		\label{fig:mnist_mnistm}
	\end{subfigure}
	\begin{subfigure}[t]{.45\textwidth}
		\centering
		\includegraphics[width=0.95\textwidth]{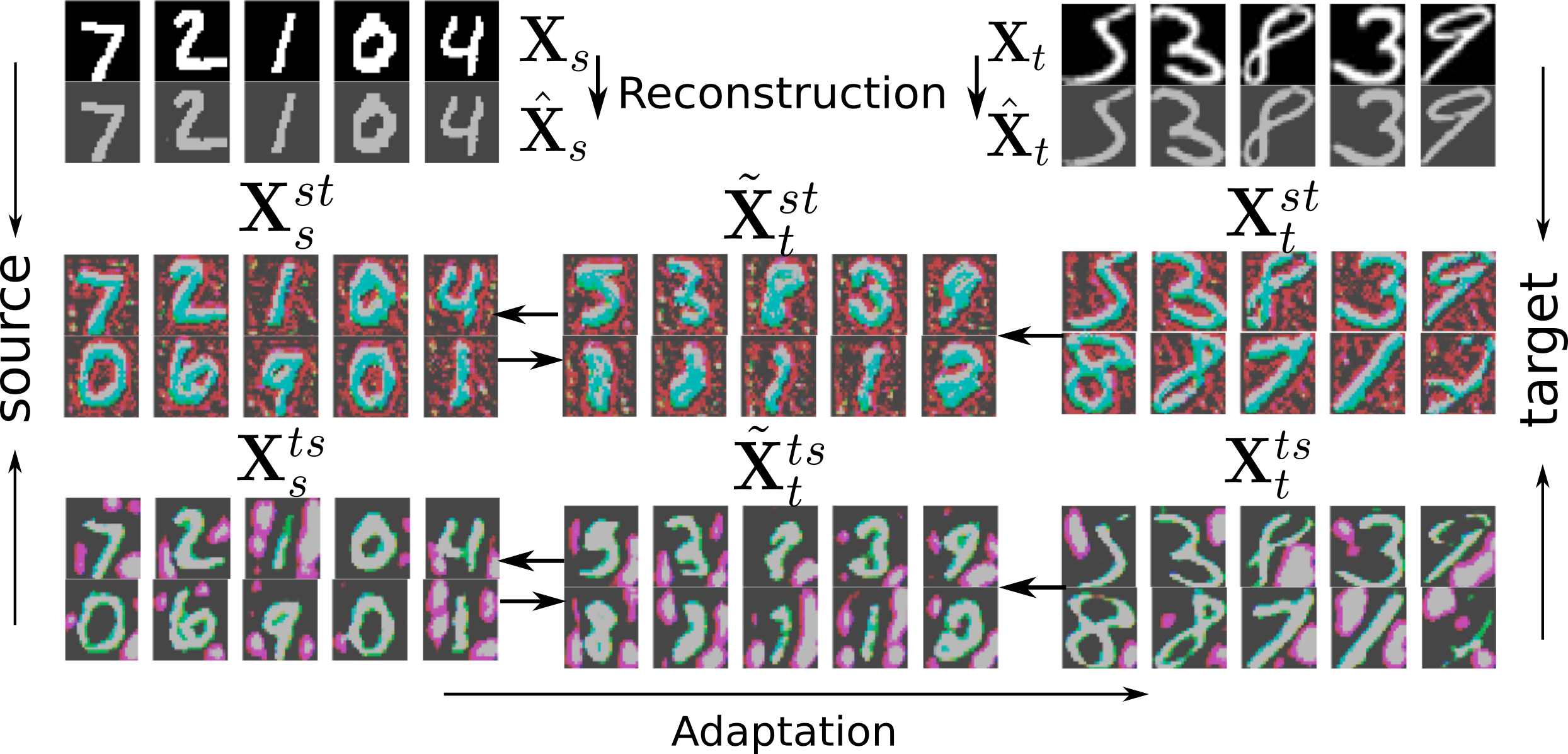}
		\caption{\small MNIST $\rightarrow$ USPS}
		\label{fig:mnist_usps}
	\end{subfigure}
	\begin{subfigure}[t]{.45\textwidth}
		\centering
		\includegraphics[width=0.95\textwidth]{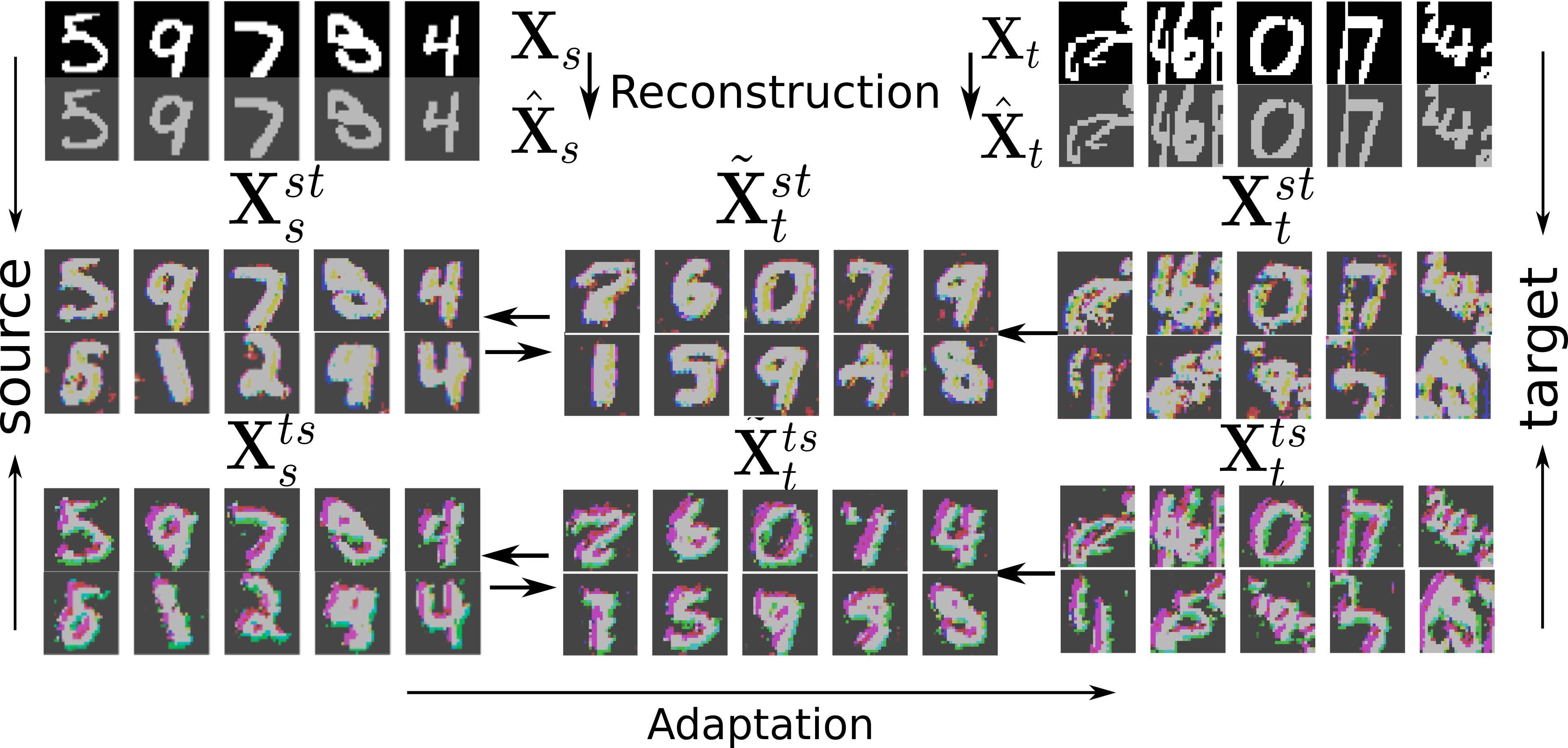}
		\caption{\small MNIST $\rightarrow$ M-Digits}
		\label{fig:mnist_multi}
	\end{subfigure}
	\begin{subfigure}[t]{.45\textwidth}
		\centering
		\includegraphics[width=0.95\textwidth]{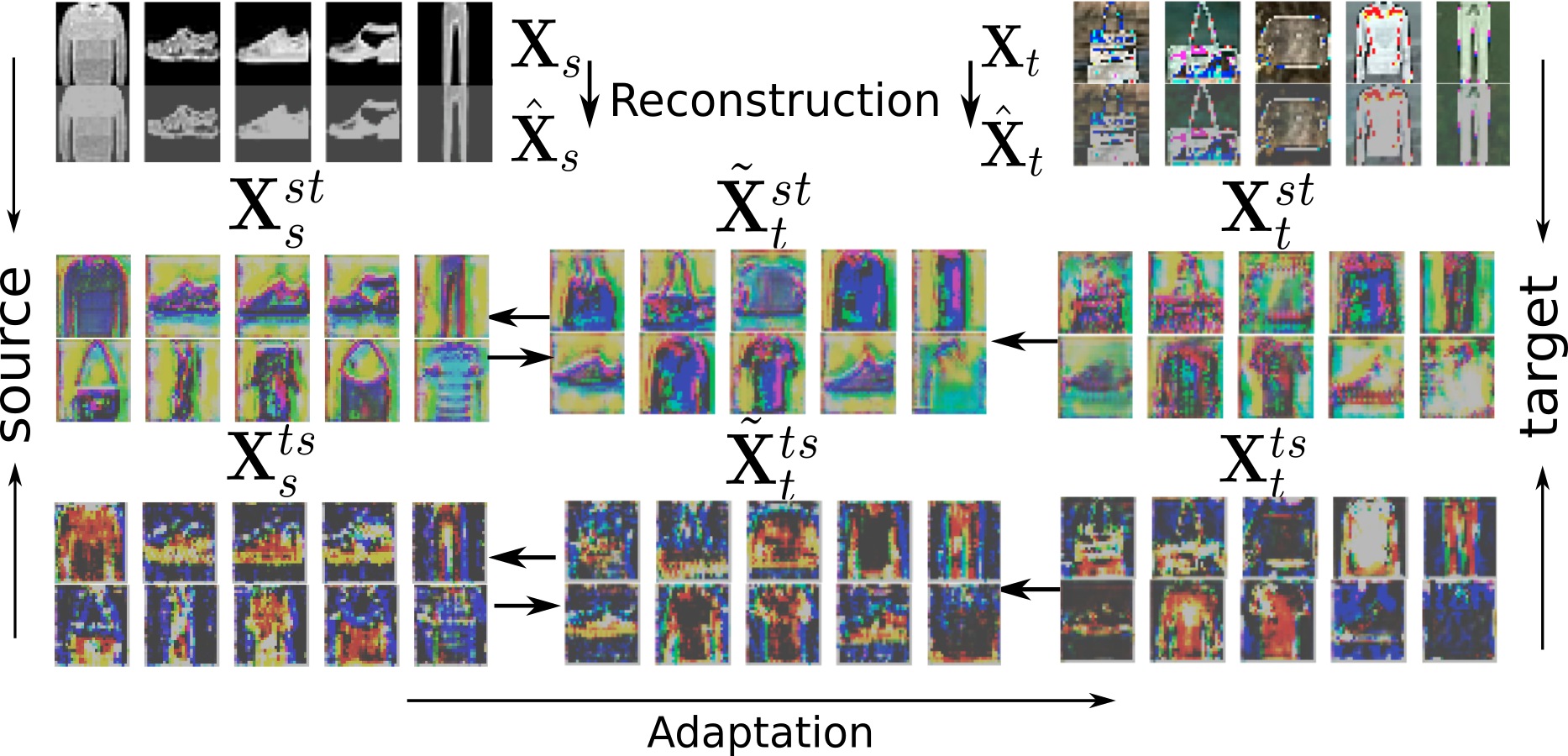}
		\caption{\small Fashion $\rightarrow$ Fashion-M}
		\label{fig:fashion_fashionm}
	\end{subfigure}
	\caption{\small The visualization of association generations. For each scenario, the leftmost column is the source and its association, and the rightmost is for target. During the experiments, the associations of source are real player. The adversarial generations for target associations are in the middle column.}
	\label{fig:img_intermediates}	
\end{figure*}

%---------------------------------------------------------------------------
\subsubsection{Qualitative Results}

Since our model adopts a generative approach, we can have direct visual evaluation of the associations generated by the CGRS. The generative associations obtained by CGRS are shown in Figure~\ref{fig:img_intermediates}, obtained after 100k mini-batch steps for the Fashion scenario and 50k for other three scenarios. The CGRS generate the associations with very similar appearance for the source and target domains. Then the GANs is employed to move them closer. During association  generation, the CGRS eliminate the strong noise of MNIST-M and Fashion-M. Though there are more complex textures in the Fashion task, the proposed model still performs well to produce reasonable visualizations of the associations. The associations of the Fashion scenario seem to suffer some information loss, possibly due to the complex textures and strongly polluted images. However, they still look reasonable upon visualization. 
%%?? 
%Also the CGRS projects the uniform background samples to the learned association space well. 
The MNIST$\rightarrow$M-Digits scenario maintains the original content style, while the associations display some style variation in the MNIST$\rightarrow$USPS scenario.

%---------------------------------------------------------------------------
\subsubsection{Model Analysis}
Some further experiments are done to evaluate our model. %In addition, we also try to find some potential advantages and limitations of our work further.
%% table for generalization
\begin{table*}
	\centering
	\caption{\small Mean classification accuracy for Generalization Evaluation. The results of $\mathbf{X}^{ts}$ channel is shown in the parentheses.}
	\begin{tabular}{|l|c|c|c|c|}
		\hline
		Source$\rightarrow$Target &  MNIST$\rightarrow$MNIST-M & MNIST$\rightarrow$USPS & MNIST$\rightarrow$M-Digits & Fashion$\rightarrow$Fashion-M\\
		%    		&  MN-M& USPS & M-Digits & F-M\\
		\hline\hline
		MNIST$\rightarrow$MNIST-M  &\textbf{0.890(0.983)} & 0.958(0.945)  & 0.915(0.853)  &  0.762(0.730) \\
		%    		\hline
		MNIST$\rightarrow$USPS &0.915(0.859)  & \textbf{0.961(0.943)} & 0.882(0.914)  & 0.605(0.587)   \\
		%    		\hline
		MNIST$\rightarrow$M-Digits &0.843(0.928) & 0.944(0.958) & \textbf{0.916(0.883)} & 0.613(0.593)  \\
		%    		\hline
		Fashion$\rightarrow$Fashion-M & 0.925(0.881) &0.932(0.935) & 0.825(0.913)  & \textbf{0.766(0.813)}  \\
		\hline
	\end{tabular}
	\label{table:generalization}
\end{table*}	

\noindent
\textbf{Ablation Study:} We evaluate the potential effect of employing the content constancy strategy in our model. From the Table~\ref{table:accuracy}, we can see that the model with content constancy (denoted by CRGS-LA-C) outperforms its CRGS-LA counterpart. The constancy loss encourages the adversarial generation in a consistent way. 

\begin{figure}
	%\centering
	\begin{subfigure}[t]{.23\textwidth}
		\centering
		\includegraphics[width=1.02\textwidth]{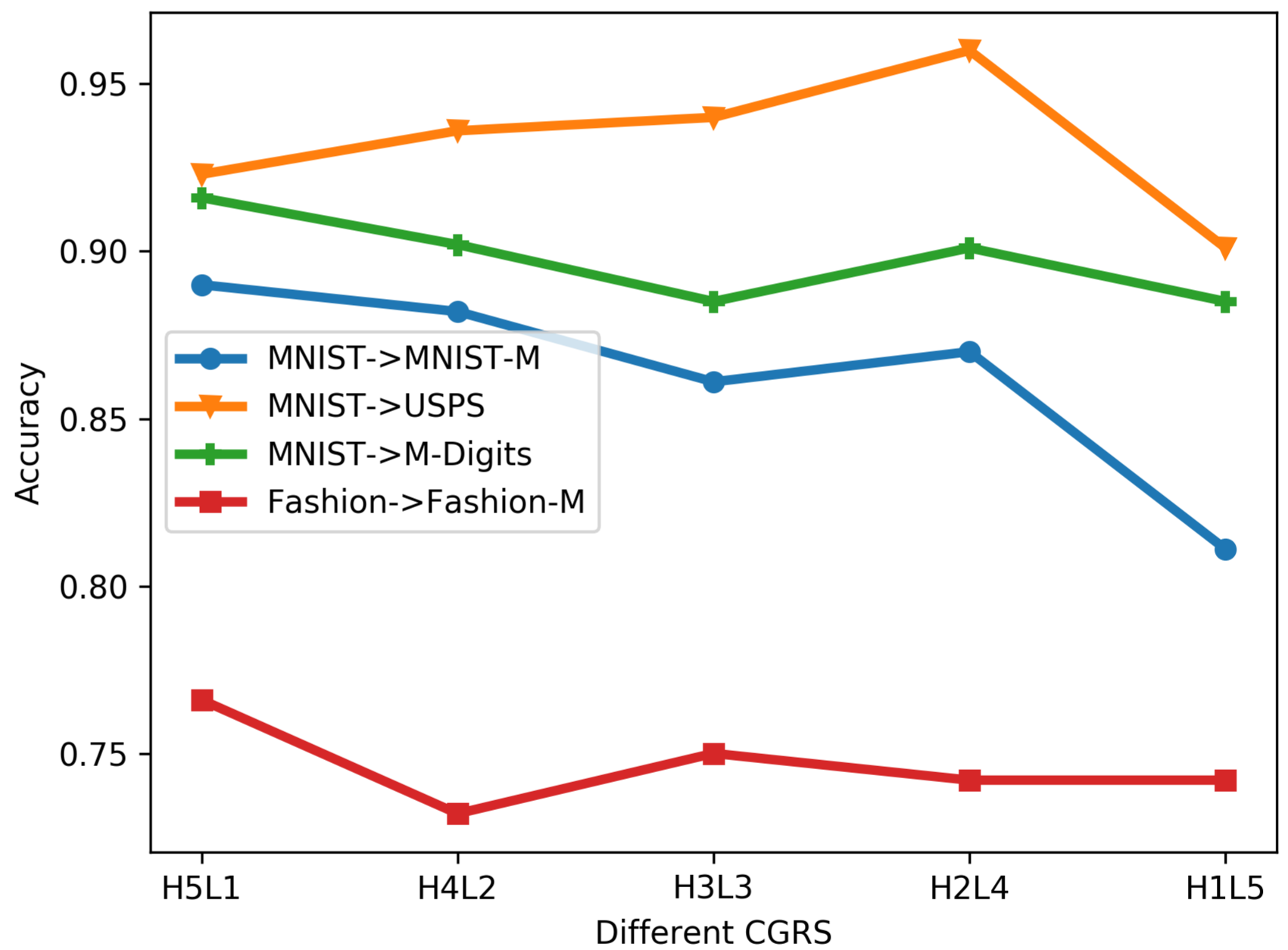}
		\caption{\small $\mathbf{X}^{st}$ channel}
		\label{fig:cgrs_st}
	\end{subfigure}
	\begin{subfigure}[t]{.23\textwidth}
		\centering
		\includegraphics[width=1.02\textwidth]{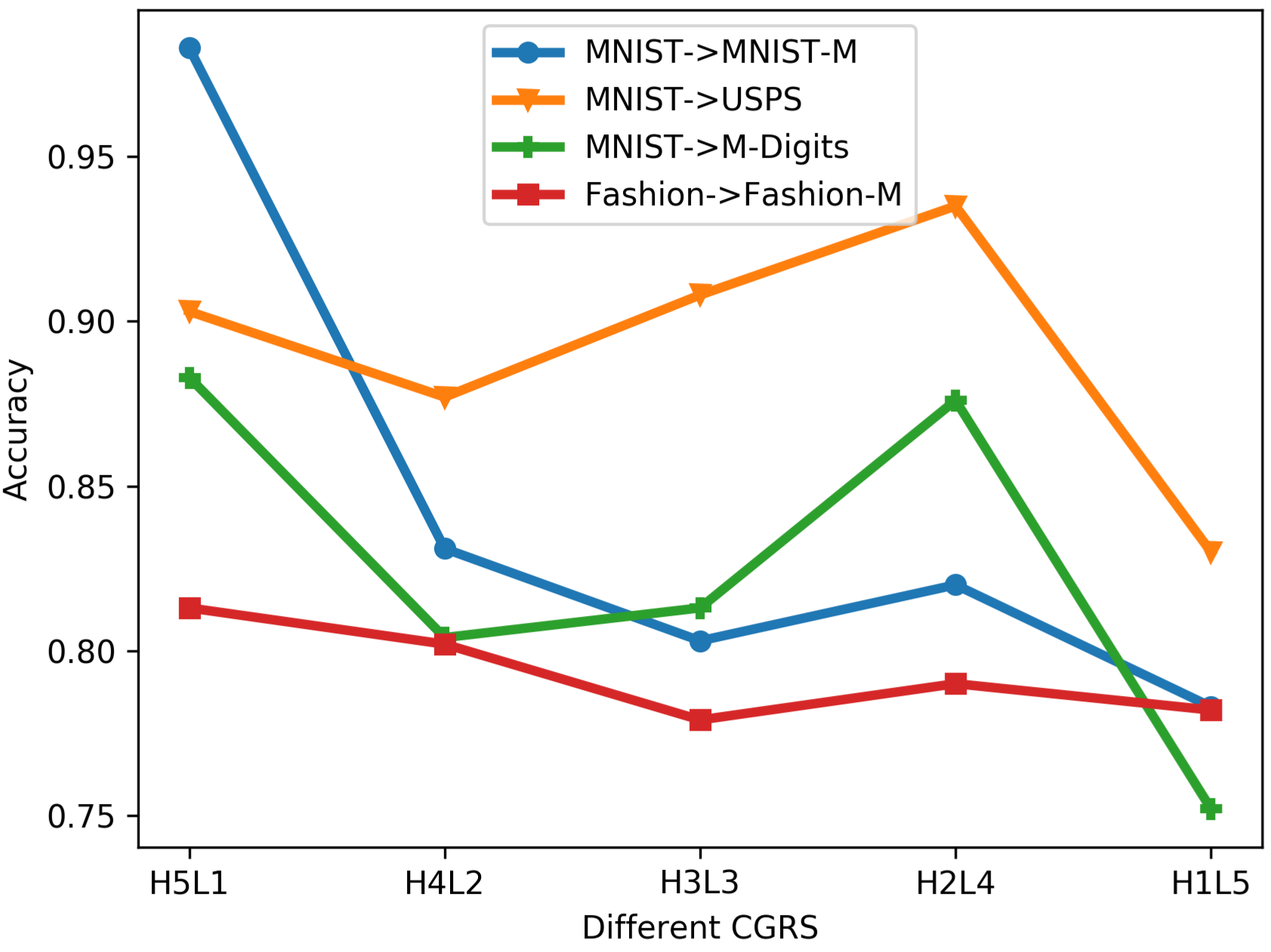}
		\caption{\small $\mathbf{X}^{ts}$ channel}
		\label{fig:cgrs_ts}
	\end{subfigure}
	\caption{\small The Adaptation Accuracy of Different CGRS.}
	\label{fig:cgrs}   	
\end{figure}

\noindent
\textbf{Sensitivity of CGRS:}  CGRS plays a critical role in the proposed model. In  this section, we evaluate the performance of diverse structures of CGRS. During the experiments, we use a fix depth of network (6 layers) for the generation process. We apply various settings for splitting the high-level and low-level decoder stacks. For example, H5L1 denotes the scheme using 5 layers for high-level and 1 layer as low-level. %And the batch normalization is added between layers. 
The results of changing the CGRS setup for different scenarios are shown in Figure~\ref{fig:cgrs}. It can be seen that for the channel $\mathbf{X}^{st}$ in MNIST$\rightarrow$MNIST-M and Fashion$\rightarrow$Fashion-M tasks, the highest accuracies are at the point H5L1, and for MNIST$\rightarrow$USPS and MNIST$\rightarrow$M-Digits tasks, there is a peak value at the point H2L4.
%% ADDED
The $\mathbf{X}^{ts}$ channel somehow seems more sensitive to varying CGRS settings.  

\noindent
\textbf{Generalization of CGRS:} Can we utilize the trained CGRS in one scenario to another adaptation task? In this evaluation, we use our pre-trained CGRS from one scenario to adapt to a different task. These models are trained with a trade-off H4L2 CGRS according to the sensitivity analysis. During the experiments, we keep the CGRS fixed, then fine-tune the adversarial and label alignment parts. The results are shown in  Table~\ref{table:generalization}. Although there is a slight reduction to the accuracies reported before, the performance of adaptation to other tasks remains reasonable. Specifically, the CGRS of MNIST$\rightarrow$MNIST-M and Fashion$\rightarrow$Fashion-M adapts to other three scenarios pretty well, while the CGRS of the MNIST$\rightarrow$USPS and MNIST$\rightarrow$M-Digits get a lower accuracy for Fashion$\rightarrow$Fashion-M.

\begin{figure}
	\centering
	\begin{subfigure}[t]{.23\textwidth}
		\centering
		\includegraphics[width=0.78\textwidth]{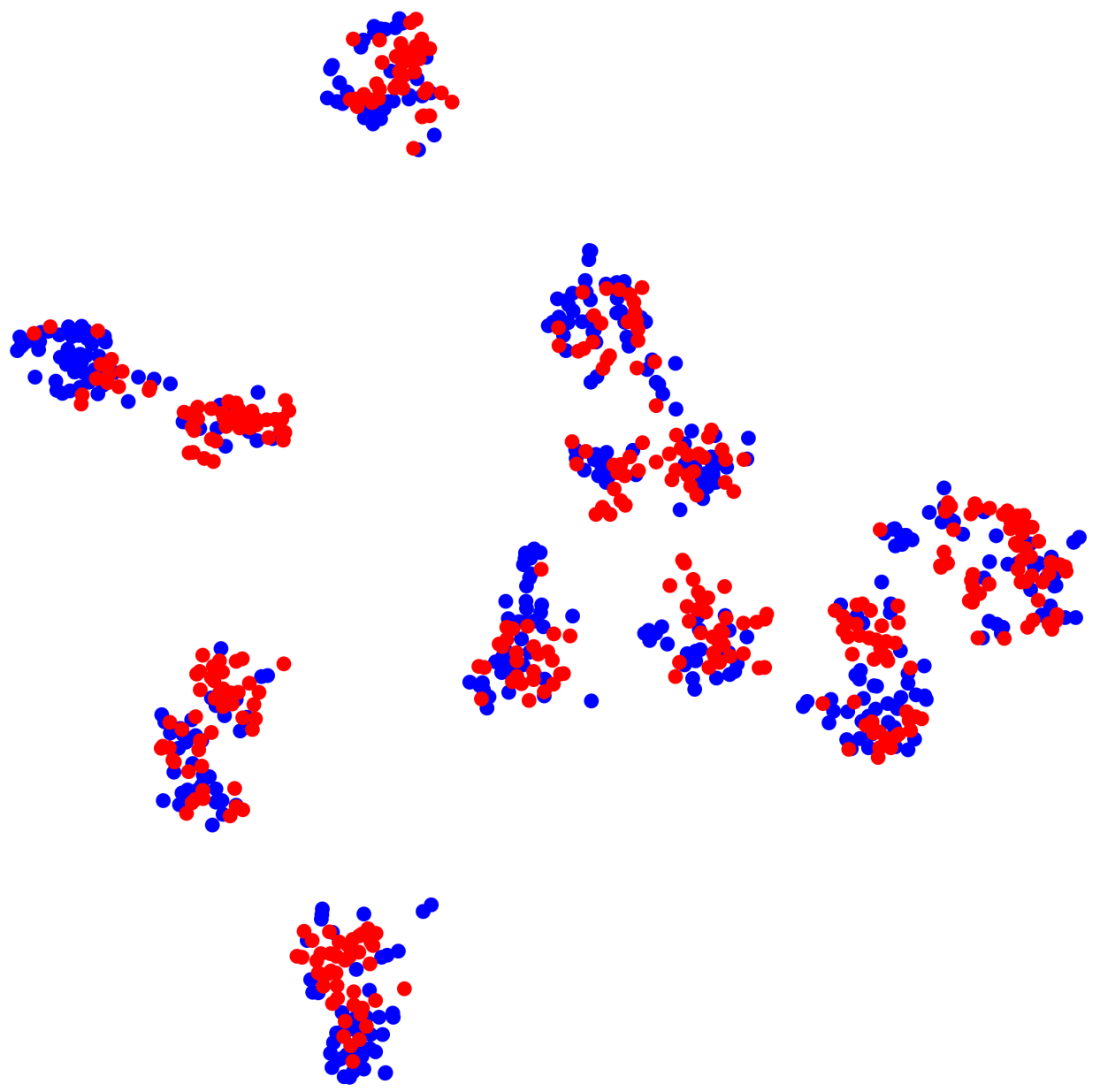}
		\caption{\small $\mathbf{X}^{st}$ channel}
		\label{fig:feat_sst}
	\end{subfigure}
	\begin{subfigure}[t]{.23\textwidth}
		\centering
		\includegraphics[width=0.78\textwidth]{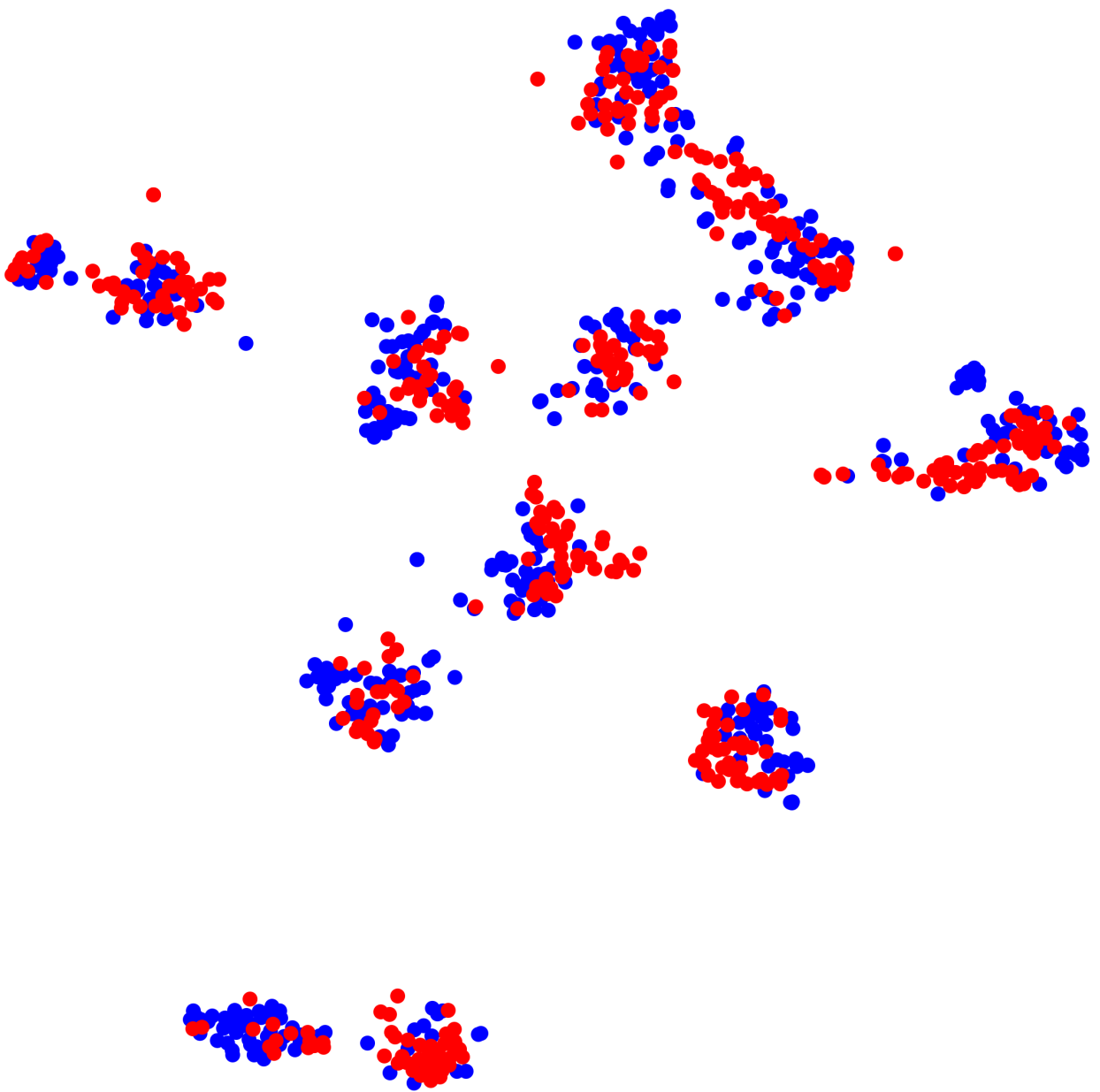}
		\caption{\small $\mathbf{X}^{ts}$ channel}
		\label{fig:feat_tst}
	\end{subfigure}
	\caption{\small The visualization of top associations features embedded by t-SNE w.r.t source and target. The Blue dots are for source and red ones for target.}
	\label{fig:feat}	
\end{figure}

\noindent
\textbf{Visualization of Extracted Features:}  We also evaluate the features of top, fully connected layers in the discriminator for task MNIST$\rightarrow$USPS. The features are embedded by the t-SNE algorithm~\cite{Maaten2008a}. Figure~\ref{fig:feat} shows that the two domains can be aligned well on both channels after adaptation.

%---------------------------------------------------------------------------
\subsubsection{Semi-supervised Evaluation}

Finally, we evaluate the performance of our model for semi-supervised learning. Under this scenario, it is assumed that we can get a small number of labeled target samples. Similar to the approach in~\cite{Bousmalis2017a}, we choose 1000 samples from every category in the target domain as the baseline. These are added as extra to the source domain for training. The results are shown in Table~\ref{table:semi-supervised}. The adaptation performance is better when some target data are added into the source to train the classifier. It outperforms the unsupervised scenario when only 1000 target samples are fed to the classifier, whereas having 2000 target samples will further improve the performance. 
\begin{table}
	\centering
	\caption{\small Mean classification accuracy for semi-supervised evaluation.}
	\begin{tabular}{|l|c|c|c|c|}
		\hline
		Source & MNIST &  MNIST & MNIST & Fashion\\
		Extra & MNIST-M & USPS & M-Digits & Fashion-M\\
		\hline\hline
		1000 & 0.988  & 0.966 & 0.925 & 0.846 \\
		2000 & 0.990 & 0.970 & 0.932  & 0.855 \\
		\hline
	\end{tabular}
	\label{table:semi-supervised}
\end{table}

\subsubsection{Discussion}
To sum up, our method can maintain stable performance when we vary the settings of CGRS for stack splitting. 
There seems to be a tendency to favour a higher ratio of high-level to low-level layers when the domains contain
similar contents but different background, while adaptation tasks with similar background but different content favour more low-level layers. 

Another interesting observation is that CGRS have very good generalization ability. The CGRS trained by one task can be employed for domain adaptation in another task. This demonstrates a merit of our method for practical applications, that is the CGRS are transferable. 

Finally, while the both association channels are well aligned, from our experiment it seems $\mathbf{X}^{ts}$ claims better classification performance more often. In practical applications, it may be possible to design a classification combination method so that an optimal final decision can be developed from both association channels. 

%--------------------------------------------------------------------------------------------------------------------------------------
\section{Conclusion}

In this paper, we have proposed a novel unsupervised domain adaptation model based on cross-domain association generation, and label alignment using adversarial networks. In particular, cross-grafted representation stacks between different domains are constructed for bi-directional associations. The domain adaptation task hence transforms to constructing an effective mapping of the cross-domain associations onto the label space of the original source domain, a methodology we believe contributes to its robust performance in domain adaptation tasks. This is verified by the empirical results we have obtained from a number of tasks involving 6 benchmark tasks, which also demonstrate that the proposed CGRS also have strong cross-task generalization abilities. For future work, we would like to explore the extension of the framework for continual learning with cross-task adaptation.

{\small
	\bibliographystyle{ieee}
	\bibliography{tl_mirror_lib}

\end{document}